\documentclass{ieeeaccess}
\usepackage{cite}
\usepackage{amsmath,amssymb,amsfonts}
\usepackage{algorithmic}
\usepackage{graphicx}
\usepackage[compatibility=false]{caption}

\usepackage{tabularx}
\usepackage{booktabs}

\def\BibTeX{{\rm B\kern-.05em{\sc i\kern-.025em b}\kern-.08em
    T\kern-.1667em\lower.7ex\hbox{E}\kern-.125emX}}
\begin{document}
\history{Date of publication xxxx 00, 0000, date of current version xxxx 00, 0000.}
\doi{10.1109/TQE.2020.DOI}

\title{Structure-Preserving Quantum Circuit Architectures for Robot Kinematics}

\author{\uppercase{Andrea Morghen}\authorrefmark{1},
\uppercase{Pierluigi Arpenti}\authorrefmark{1},
\uppercase{Roberto Schiattarella}\authorrefmark{2},
\uppercase{Giovanni Acampora}\authorrefmark{2},
and \uppercase{Bruno Siciliano}\authorrefmark{1}}

\address[1]{Department of Electrical Engineering and Information Technology (DIETI), University of Naples Federico II, 80125 Naples, Italy (e-mail: andrea.morghen@unina.it; pierluigi.arpenti@unina.it; bruno.siciliano@unina.it)}

\address[2]{Department of Physics ``Ettore Pancini'', University of Naples Federico II, 80126 Naples, Italy (e-mail: roberto.schiattarella@unina.it; giovanni.acampora@unina.it)}

\markboth
{Morghen \headeretal: Structure-Preserving Quantum Circuit Architectures for Robot Kinematics}
{Morghen \headeretal: Structure-Preserving Quantum Circuit Architectures for Robot Kinematics}

\corresp{Corresponding author: Andrea Morghen (e-mail: andrea.morghen@unina.it).}

\begin{abstract}
Structured spatial data require quantum encodings that preserve geometric relations, expose measurable observables, and remain implementable on finite-depth hardware. This work introduces a quantum representation and circuit architecture for rigid-body transformations and specializes it to Denavit--Hartenberg kinematics of serial open-chain manipulators. Each translational contribution is factorized into a classical metric magnitude and a signed unit direction encoded by a single-qubit Bloch vector, while parameterized rotations reproduce the ordered propagation of frame directions. A selector register prepares probabilities proportional to the contribution magnitudes, and the reduced state of a designated readout qubit encodes their normalized weighted sum. The retained classical scale then reconstructs the metric end-effector position. Two additional readout qubits encode terminal-frame axes, providing a compact and geometrically interpretable pose interface. At the ideal expectation-value level, measured Pauli observables reproduce the corresponding classical kinematic quantities. Alternative circuit architectures realize the same representation with different tradeoffs in qubit count, circuit depth, controlled operations, and measurement requirements. Validation on a serial manipulator yields numerically negligible position and orientation reconstruction errors under ideal simulation. Finite-shot simulations, noisy executions, transpilation analysis, and a hardware demonstration further characterize statistical error, noise sensitivity, and implementation overhead without asserting computational advantage.
\end{abstract}


\titlepgskip=-15pt


\maketitle

\section{Introduction}
\label{sec:introduction}

\IEEEPARstart{Q}{uantum} engineering requires mappings that translate structured classical data into quantum states, circuit parameters, and measurable observables while preserving the relations that define the original problem. This requirement is particularly relevant for rigid-body transformations, which combine spatial rotations, metric translations, and ordered frame composition. Although unit directions can be naturally represented by qubit Bloch vectors and propagated through the correspondence between $\mathrm{SU}(2)$ and $\mathrm{SO}(3)$, non-unit translations and weighted sums of spatial contributions require explicit preparation, aggregation, measurement, and reconstruction procedures.
Serial-manipulator kinematics provides a structured application in which these issues arise naturally. Under the Denavit--Hartenberg (DH) convention, the end-effector pose is generated through an ordered sequence of joint-dependent rotations and link translations \cite{Denavit1955KinematicNotation,Craig2005IntroductionRobotics,Siciliano2009Robotics}. The position can be expressed as a weighted sum of configuration-dependent directions, whereas the orientation is determined by the propagated terminal-frame axes. A quantum representation of this model should therefore preserve the semantics of the joint variables, the ordered propagation of frame directions, and the distinction between directional and metric information.
A growing body of preliminary research has explored quantum and hybrid quantum--classical methods for robotic navigation, planning, coordination, learning, posture representation, and kinematic computation \cite{Yan2024QuantumRobotics,Hohenfeld2024QuantumNavigation,Chella2023QuantumSwarm,Mannone2025QuantumSwarm,Zioui2021Orientation,Fazilat2022ForwardKinematics,Otani2025Posture}. These studies demonstrate increasing interest in quantum robotics and provide initial evidence that qubit states and observables can represent selected robotic quantities. However, they do not yet establish a general structure-preserving circuit formulation that maps a standard robot description to prepared quantum states, preserves the full kinematic construction, and provides explicit position-and-orientation estimators with quantifiable resource and measurement costs.

This work addresses this gap by introducing an executable circuit-level mapping for structured rigid-body transformations and specializing it to standard and Craig modified-DH kinematics. Each translational contribution is decomposed into a classical metric magnitude and a signed unit direction. The direction is represented by a qubit Bloch vector and prepared through parameterized single-qubit rotations that reproduce the ordered propagation of the corresponding frame axis. The sign is incorporated into the prepared direction, while the nonnegative magnitudes determine the aggregation probabilities. Dedicated qubits similarly encode selected terminal-frame axes for orientation readout.

The central circuit problem is the aggregation of the translational contributions. Because arbitrary quantum states cannot be combined through direct vector addition, the proposed compact architectures use a selector register whose branch probabilities are proportional to the contribution magnitudes. Conditional operations associate each branch with the corresponding directional state, such that the reduced state of a designated position qubit encodes the normalized weighted sum. The retained classical scale factor then reconstructs the metric end-effector position from the measured Pauli expectation values.
Two compact circuit architectures are developed. The Item-Bank Selector-SWAP (IBSS) architecture prepares the complete bank of directional item states and routes the selected state to the position qubit through selector-controlled SWAP operations. The Selector-Controlled Preparation (SCP) architecture eliminates the explicit item bank and directly prepares the selected direction on the position qubit through selector-controlled rotations. The two constructions produce the same ideal reduced position state but distribute resources differently between item qubits, selector preparation, controlled operations, circuit depth, and routing overhead.
The compact architectures are complemented by three NISQ-oriented alternatives. Item-Wise Readout (IWR) avoids quantum aggregation by measuring the directional contributions separately and reconstructing their weighted sum classically, exchanging a larger measurement cost for shallower circuits and fewer multi-qubit controls. Tree-factorized variants replace flat selector structures with hierarchical lower-fan-in operations to redistribute control depth and routing requirements. Parallel XYZ readout assigns Cartesian observables to distinct probe qubits, reducing the number of measurement settings at the cost of additional qubits and replicated preparation operations. These alternatives define a circuit-design space in which width, depth, controlled-operation count, measurement settings, circuit executions, and classical post-processing can be traded according to the target processor and application.
The main contributions of this work are as follows:
\begin{itemize}
\item An executable circuit-level mapping from structured rigid-body transformations to parameterized quantum operations, directional states, and measurable pose estimators, with a structure-preserving specialization to serial-manipulator DH kinematics.

\item Two compact aggregation architectures, IBSS and SCP, together with an ideal reduced-state equivalence showing that both encode the same normalized weighted position vector.

\item Three NISQ-oriented alternatives---IWR, tree-factorized selectors, and parallel XYZ readout---that expose explicit trade-offs among circuit width, controlled-operation depth, measurement settings, circuit executions, and classical reconstruction.

\item A resource analysis that distinguishes logical, decomposed, transpiled, and executed circuits and reports qubits, ancillary and selector registers, circuit depth, one- and two-qubit gates, controlled operations, measurement settings, shots, circuit executions, and classical processing.

\item An experimental characterization under ideal statevector, finite-shot, and noisy-simulation conditions, followed by execution on a real IBM quantum processor.

\item A physical-robot validation in which the pose reconstructed from the quantum-circuit outputs is compared with the corresponding pose of a real serial manipulator, providing a proof of implementability across the quantum and robotic components of the experimental pipeline.
\end{itemize}
The proposed architectures are not intended to replace classical forward kinematics, and no claim of quantum speedup or computational advantage is made. Rather, they provide a measurable and executable quantum representation of robot kinematics parameterized by the physical joint variables. This representation can serve as a kinematic module within fully quantum or hybrid quantum--classical architectures, allowing downstream quantum operations to access pose-related observables without replacing the underlying joint-space semantics with an unrelated latent parameterization. The architectural comparison therefore quantifies how different circuit realizations of this interface affect resource requirements, statistical uncertainty, noise sensitivity, and hardware feasibility.

\begin{table*}[!t]
\centering
\caption{Comparison with representative quantum representations of robotic geometry and kinematics.}
\label{tab:related_work_comparison}
\renewcommand{\arraystretch}{1.12}
\setlength{\tabcolsep}{3.0pt}
\scriptsize
\begin{tabularx}{\textwidth}{
@{}
>{\raggedright\arraybackslash}p{2.45cm}
>{\raggedright\arraybackslash}p{2.85cm}
>{\raggedright\arraybackslash}p{3.75cm}
>{\raggedright\arraybackslash}p{4.05cm}
>{\raggedright\arraybackslash}X
@{}
}
\toprule
\textbf{Work}
&
\textbf{Represented quantity}
&
\textbf{Model construction}
&
\textbf{Translation and pose readout}
&
\textbf{Circuit scope}
\\
\midrule

Zioui et al.~\cite{Zioui2021Orientation}
&
Rigid-body and manipulator orientation
&
Geometric single-qubit rotation mapping
&
Orientation from Pauli expectation values; translation not addressed
&
Orientation-level realization
\\

Fazilat et al.~\cite{Fazilat2022ForwardKinematics}
&
Robot pose
&
Robot-specific quaternion--Pauli construction derived from DH data
&
Quantum rotational processing with classical translational composition
&
Single quantum--classical realization
\\

Otani et al.~\cite{Otani2025Posture}
&
Link posture and end-effector position
&
Parameterized link-direction states
&
Item-wise Pauli readout with classical length weighting and summation
&
No compact position-state aggregation
\\

Nigatu et al.~\cite{Nigatu2025QMLGrover}
&
Forward-kinematics surrogate
&
Data-trained parameterized quantum circuit
&
Pose approximated by the learned model
&
Surrogate rather than analytic circuit synthesis
\\

\textbf{This work}
&
\textbf{Rigid-body pose and serial-manipulator kinematics}
&
\textbf{Circuit synthesis from structured transformations and DH data}
&
\textbf{Compact or item-wise position reconstruction and terminal-frame directional readout}
&
\textbf{Alternative compact and NISQ-oriented architectures with resource and execution analysis}
\\

\bottomrule
\end{tabularx}
\end{table*}

\section{Related Work}
\label{sec:related_work}

Research on quantum robotics is still preliminary but has expanded across manipulation, navigation, planning, coordination, learning, and optimization \cite{Yan2024QuantumRobotics,Hohenfeld2024QuantumNavigation,Chella2023QuantumSwarm,Mannone2025QuantumSwarm}. Most existing studies introduce quantum computation at the task or solver level, for example by encoding candidate solutions, discretized joint configurations, or optimization objectives. Although these contributions demonstrate growing interest in quantum methods for robotics, they do not generally represent the continuous kinematic construction of a robot within the quantum circuit. Since the present work concerns the circuit realization of structured rigid-body data the following discussion focuses on approaches that directly encode geometric or kinematic quantities.

\subsection{Quantum Representations of Robotic Geometry}
\label{subsec:rw_quantum_geometry}

Zioui et al.\ related rigid-body and robotic-arm orientations to single-qubit states by exploiting the correspondence among Bloch vectors, Pauli observables, and spatial rotations \cite{Zioui2021Orientation}. Their formulation establishes an interpretable quantum representation of orientation, but it does not address translational composition or the complete pose of a general serial chain.\\
Fazilat et al.\ considered the forward kinematics of an ABB IRB~140 manipulator through quaternion--Pauli correspondences derived from its Denavit--Hartenberg description \cite{Fazilat2022ForwardKinematics}. Rotational quantities are processed through quantum operations, whereas the translational component is assembled through classical arithmetic. The approach therefore provides a robot-specific quantum--classical realization rather than a general circuit construction for the quantum aggregation and readout of weighted translational directions.\\
Otani et al.\ represented robot-link directions through parameterized qubit states and reconstructed the end-effector position by weighting Pauli expectation values with classical link lengths \cite{Otani2025Posture}. This work demonstrates that link-level directional states can retain an interpretable relationship with robot posture. However, the circuit is not systematically synthesized from a standard or modified Denavit--Hartenberg model, and no explicit one-to-one correspondence is derived or validated between the circuit parameters and the physical joint variables of the manipulator. Consequently, the parameterized rotations cannot generally be interpreted as preserving the robot joint-coordinate semantics. Moreover, the directional contributions are measured individually and combined through classical post-processing rather than aggregated into a compact position state.\\
Nigatu et al.\ used a parameterized quantum circuit as a learned surrogate for forward kinematics within a quantum-assisted search procedure \cite{Nigatu2025QMLGrover}. This data-driven approach differs from an analytic construction in which circuit parameters and operations are derived directly from the physical joint variables and rigid transformations of the robot model.

These studies provide initial representations of orientations, link directions, or robot-specific pose quantities. However, the reviewed literature does not provide, within a unified model-derived framework, both compact and item-wise circuit realizations of the same position-and-orientation target, together with an explicit separation between classical metric magnitudes and quantum directional states.

\subsection{Positioning of the Present Work}
\label{subsec:rw_positioning}

The present work addresses this gap at the circuit-architecture level. Starting from structured rigid-body and Denavit--Hartenberg data, it constructs parameterized quantum circuits that preserve the distinction between classical metric quantities and quantum directional states. It then compares alternative realizations of the same ideal kinematic target, including compact selector-based architectures and NISQ-oriented variants based on item-wise reconstruction, tree-factorized controls, and parallel Cartesian readout. The comparison therefore considers not only pose reconstruction accuracy, but also logical width, circuit depth, controlled operations, measurement settings, repeated executions, and classical post-processing.\\
The proposed constructions draw on established quantum-engineering primitives for probability-amplitude preparation, uniformly controlled rotations, quantum multiplexing, and general circuit synthesis \cite{Bergholm2005UniformlyControlled,Plesch2011StatePreparation,Shende2006CircuitSynthesis}. These primitives do not by themselves define how the spatial structure of a serial manipulator should be translated into circuit parameters, register organizations, aggregation procedures, and measurable pose estimators. This distinction is particularly relevant for noisy intermediate-scale quantum processors, where limited coherence, imperfect gates, restricted connectivity, and finite sampling can make logically compact circuits costly after decomposition, routing, and transpilation \cite{Preskill2018NISQ,Bharti2022NISQAlgorithms}. The proposed architectures are therefore evaluated at the logical, decomposed, transpiled, finite-shot, noisy, and hardware-execution levels.\\

\section{Methodological Background}
\label{sec:methodological_background}

This section introduces the mathematical and quantum-mechanical concepts required to formulate the proposed circuit architectures. We first define the representation and composition of rigid-body transformations and the directional decomposition induced by serial-manipulator kinematics. We then introduce directional qubit states, their evolution under single-qubit rotations, selector-controlled operations, reduced-state readout, and finite-shot Pauli estimation.

\subsection{Rigid-Body Transformations and Directional Decomposition}
\label{subsec:robot_model}

A rigid-body transformation is represented by
\[
\mathbf T
=
(\mathbf R,\mathbf p)
=
\begin{bmatrix}
\mathbf R & \mathbf p\\
\mathbf 0_3^{\mathsf T} & 1
\end{bmatrix}
\in SE(3),
\]
where $\mathbf R\in SO(3)$ is a rotation matrix and $\mathbf p\in\mathbb R^3$ is a position vector. Given two transformations $\mathbf T=(\mathbf R,\mathbf p)$ and $\mathbf T'=(\mathbf R',\mathbf p')$, their composition is
$\mathbf T\mathbf T' = \left(\mathbf R\mathbf R',\mathbf p+\mathbf R\mathbf p'\right)$.
Thus, rotational components compose through matrix multiplication, whereas each translational component is first expressed in the preceding reference frame and then added to the accumulated position.
Consider a serial open-chain manipulator with $n\in\mathbb N$ one-degree-of-freedom joint coordinates and configuration $\mathbf q = [q_1,\ldots,q_n]^{\mathsf T} \in \mathcal Q = \prod_{i=1}^{n}\mathcal Q_i \subseteq \mathbb R^n$, where $\mathcal Q_i$ is the admissible set of the $i$-th joint coordinate. Each $q_i$ describes either a revolute or a prismatic elementary motion.
Let $\mathcal F_B$ denote the fixed reference frame and let $\mathcal F_0,\ldots,\mathcal F_n$ denote the frames attached to the kinematic chain, with $\mathcal F_n$ identified with the end-effector frame. Throughout this article, $\mathbf T_i^{j}$ denotes the pose of frame $\mathcal F_i$ expressed in frame $\mathcal F_j$ and when a transformation, rotation, position, or spatial direction is expressed in the base frame $\mathcal F_B$, the superscript $B$ is omitted for compactness. Accordingly, the constant pose of $\mathcal F_0$ expressed in the base frame is $\mathbf T_i \equiv \mathbf T_0^{B} = ({\mathbf R}_0,{\mathbf p}_0)$.
The relative transformation describing the pose of $\mathcal F_i$ expressed in the preceding frame $\mathcal F_{i-1}$ is denoted by
$\mathbf T_i^{i-1}(q_i) = \left( \mathbf R_i^{i-1}(q_i), \mathbf p_i^{i-1}(q_i) \right)$. 
The forward-kinematics map $\mathbf T_n: \mathcal Q \rightarrow SE(3)$ is therefore
\begin{equation}
\mathbf T_n(\mathbf q)
=
\mathbf T_0
\prod_{i=1}^{n}
\mathbf T_i^{i-1}(q_i)
=
\begin{bmatrix}
\mathbf R_n(\mathbf q) & \mathbf p_n(\mathbf q)\\
\mathbf 0_3^{\mathsf T} & 1
\end{bmatrix},
\label{eq:forward_kinematic_transform}
\end{equation}
where the product is ordered as $\mathbf T_1^{0}(q_1) \mathbf T_2^{1}(q_2) \cdots \mathbf T_n^{n-1}(q_n)$. The maps $\mathbf R_n:\mathcal Q\to SO(3)$ and $\mathbf p_n:\mathcal Q\to\mathbb R^3$ denote the end-effector orientation and position expressed in $\mathcal F_B$.\\
Under the standard Denavit--Hartenberg (SDH) and Craig modified-Denavit--Hartenberg (MDH) conventions, the \(i\)-th rows of the corresponding parameter tables are
\begin{equation*}
    \begin{aligned}
         \mathcal D_i^{\mathrm{SDH}}(q_i)&=\bigl(a_i,\alpha_i,d_i(q_i),\theta_i(q_i)\bigr),\\
         \mathcal D_i^{\mathrm{MDH}}(q_i)&=\bigl(a_{i-1},\alpha_{i-1},d_i(q_i),\theta_i(q_i)\bigr).
    \end{aligned}
\end{equation*}
for \(i=1,\ldots,n\). For a revolute joint $\theta_i(q_i)=\bar{\theta}_i+q_i$ and $d_i(q_i)=\bar d_i$, whereas, for a prismatic joint $d_i(q_i)=\bar d_i+q_i$ and $\theta_i(q_i)=\bar{\theta}_i$.
The quantities \(a_i\), \(\alpha_i\), \(\bar d_i\), and \(\bar{\theta}_i\) are classical geometric parameters, while \(\mathbf q\) provides the configuration-dependent classical input from which the circuit rotation angles and translational weights are derived.
For $\nu\in\{x,y,z\}$, define $\operatorname{Rot}_{\nu}(\phi)=\mathbf T\!\left(\mathbf R_{\nu}(\phi),\mathbf 0_3\right)\in SE(3)$ and $\operatorname{Tra}_{\nu}(\ell)=\mathbf T\!\left(\mathbf I_3,\ell\mathbf e_{\nu}\right)\in SE(3)$, where $\mathbf R_{\nu}(\phi)\in SO(3)$ is the right-handed rotation matrix of angle $\phi$ about the $\nu$-axis and $\ell\mathbf e_{\nu}\in\mathbb R^3$ is the translation vector of length $\ell$ along the same axis. The SDH and MDH row transformations \cite{Denavit1955KinematicNotation,Craig2005IntroductionRobotics} are
\[
\begin{aligned}
\mathbf T_{i,\mathrm{SDH}}^{i-1}(q_i)
&=
\operatorname{Rot}_z\!\bigl(\theta_i\bigr)
\operatorname{Tra}_z\!\bigl(d_i\bigr)
\operatorname{Tra}_x(a_i)
\operatorname{Rot}_x(\alpha_i),
\\
\mathbf T_{i,\mathrm{MDH}}^{i-1}(q_i)
&=
\operatorname{Rot}_x(\alpha_{i-1})
\operatorname{Tra}_x(a_{i-1})
\operatorname{Rot}_z\!\bigl(\theta_i\bigr)
\operatorname{Tra}_z\!\bigl(d_i\bigr).
\end{aligned}
\] Hereafter, the dependence of \(d_i\) and \(\theta_i\) on \(q_i\) is omitted when unambiguous. Let \(\mathbf e_x,\mathbf e_y,\mathbf e_z\in\mathbb S^2\) denote the canonical Cartesian unit vectors, where $\mathbb S^2 = \left\{ \mathbf v\in\mathbb R^3  \mid \|\mathbf  \|_2=1\right\}$. Define \(\mathbf q_{1,i}:=[q_1,\ldots,q_i]^{\mathsf T}\), using the composition rule in \eqref{eq:forward_kinematic_transform}, the SDH recursion is
\begin{equation}
\begin{aligned}
\mathbf R_i(\mathbf q_{1,i})
&=
\mathbf R_{i-1}(\mathbf q_{1,i-1})
\mathbf R_z\!\bigl(\theta_i(q_i)\bigr)
\mathbf R_x(\alpha_i),
\\
\mathbf p_i(\mathbf q_{1,i})
&=
\mathbf p_{i-1}(\mathbf q_{1,i-1})
+
d_i(q_i)\,
\mathbf R_{i-1}(\mathbf q_{1,i-1})\mathbf e_z
\\
&\quad+
a_i\,
\mathbf R_{i-1}(\mathbf q_{1,i-1})
\mathbf R_z\!\bigl(\theta_i(q_i)\bigr)\mathbf e_x,
\end{aligned}
\label{eq:sdh_recursion}
\end{equation}
so the end-effector position is 
\[
\mathbf p_n(\mathbf q)
=
{\mathbf p}_0
+
\sum_{i=1}^{n}
d_i\,
\mathbf u_{i,d}^{\mathrm{SDH}}(\mathbf q_{1,i-1})
+
\sum_{i=1}^{n}
a_i\,
\mathbf u_{i,a}^{\mathrm{SDH}}(\mathbf q_{1,i}),
\]
where $\mathbf u_{i,d}^{\mathrm{SDH}}(\mathbf q_{1,i-1}) = \mathbf R_{i-1}(\mathbf q_{1,i-1})\mathbf e_z,$ and $\mathbf u_{i,a}^{\mathrm{SDH}}(\mathbf q_{1,i}) = \mathbf R_{i-1}(\mathbf q_{1,i-1}) \mathbf R_z\!\bigl(\theta_i(q_i)\bigr)\mathbf e_x$ are the propagated SDH translation directions. Whereas the MDH recursion is
\begin{equation}
\begin{aligned}
\mathbf R_i(\mathbf q_{1,i})
&=
\mathbf R_{i-1}(\mathbf q_{1,i-1})
\mathbf R_x(\alpha_{i-1})
\mathbf R_z\!\bigl(\theta_i(q_i)\bigr),
\\
\mathbf p_i(\mathbf q_{1,i})
&=
\mathbf p_{i-1}(\mathbf q_{1,i-1})
+
a_{i-1}\,
\mathbf R_{i-1}(\mathbf q_{1,i-1})\mathbf e_x
\\
&\quad+
d_i(q_i)\,
\mathbf R_{i-1}(\mathbf q_{1,i-1})
\mathbf R_x(\alpha_{i-1})\mathbf e_z,
\end{aligned}
\label{eq:mdh_recursion}
\end{equation}
hence, the end-effector position is 
\[
\mathbf p_n(\mathbf q) \hspace{-0.05cm}
=\hspace{-0.05cm}
{\mathbf p}_0 \hspace{-0.05cm}
+\hspace{-0.2cm}
\sum_{i=1}^{n}
a_{i-1}
\mathbf u_{i,a}^{\mathrm{MDH}}(\mathbf q_{1,i-1}) \hspace{-0.05cm}
+\hspace{-0.2cm}
\sum_{i=1}^{n}
d_i
\mathbf u_{i,d}^{\mathrm{MDH}}(\mathbf q_{1,i-1}),
\]
where $\mathbf u_{i,d}^{\mathrm{MDH}}(\mathbf q_{1,i-1}) = \mathbf R_{i-1}(\mathbf q_{1,i-1}) \mathbf R_x(\alpha_{i-1})\mathbf e_z$  and $\mathbf u_{i,a}^{\mathrm{MDH}}(\mathbf q_{1,i-1}) = \mathbf R_{i-1}(\mathbf q_{1,i-1})\mathbf e_x$ are the propagated SDH translation. 
Both DH conventions admit the same item-wise representation. Let $\mathcal I$ denote the fixed set of translational items induced by the DH model whose metric contribution is nonzero for at least one configuration $\mathbf q\in\mathcal Q$, and let $K=|\mathcal I|$. For each $k\in\mathcal I$ and configuration $\mathbf q\in\mathcal Q$, define the kinematic item $\eta_k(\mathbf q)=(\ell_k(\mathbf q),\mathbf s_k(\mathbf q),\mathbf A_k(\mathbf q))$, where $\ell_k(\mathbf q)\geq0$ is the absolute magnitude of the corresponding DH translational coefficient, $\mathbf s_k(\mathbf q)\in\{\pm\mathbf e_x,\pm\mathbf e_z\}$ is the associated local translation axis with the sign of that coefficient absorbed into its orientation, and $\mathbf A_k(\mathbf q)\in SO(3)$ is the rotation prefix that expresses the direction in the base frame. The propagated direction is $\mathbf u_k(\mathbf q)=\mathbf A_k(\mathbf q)\mathbf s_k(\mathbf q)\in\mathbb S^2$, so that the Cartesian contribution of item $k$ is $\ell_k(\mathbf q)\mathbf u_k(\mathbf q)$. Items whose metric magnitude vanishes at a particular configuration remain part of $\mathcal I$ but contribute zero to the translational sum. To keep $\mathbf s_k(\mathbf q)$ unit-norm also when the corresponding metric coefficient vanishes, we use the convention $\operatorname{sgn}_{+}(c)=+1$ for $c\geq0$ and $\operatorname{sgn}_{+}(c)=-1$ for $c<0$; the choice at $c=0$ is immaterial because the associated magnitude $\ell_k(\mathbf q)$ is zero.Table~\ref{tab:dh_candidate_items} specifies the item quantities generated by each DH row. For compactness, it uses $\mathbf R_{i-1}\equiv\mathbf R_{i-1}(\mathbf q_{1,i-1})$.
\begin{table}[!t]
\caption{Translational items induced by the \(i\)-th DH row when the corresponding metric coefficient is nonzero.}
\label{tab:dh_candidate_items}
\centering
\footnotesize
\setlength{\tabcolsep}{2.6pt}
\renewcommand{\arraystretch}{1.15}
\begin{tabularx}{\columnwidth}{
@{}
>{\centering\arraybackslash}p{0.10\columnwidth}
>{\centering\arraybackslash}p{0.10\columnwidth}
>{\centering\arraybackslash}p{0.15\columnwidth}
>{\centering\arraybackslash}p{0.26\columnwidth}
>{\raggedright\arraybackslash}X
@{}
}
\toprule
\textbf{DH} & \textbf{Item} & \(\boldsymbol{\ell_k(\mathbf q)}\) & \(\boldsymbol{\mathbf s_k(\mathbf q)}\) & \(\boldsymbol{\mathbf A_k(\mathbf q)}\) \\
\midrule
SDH & \(\eta_{i,d}\) & \(\left|d_i(q_i)\right|\) & \(\operatorname{sgn}_{+}\!\left(d_i(q_i)\right)\mathbf e_z\) & \(\mathbf R_{i-1}\) \\
SDH & \(\eta_{i,a}\) & \(\left|a_i\right|\) & \(\operatorname{sgn}_{+}(a_i)\mathbf e_x\) & \(\mathbf R_{i-1}\mathbf R_z\!\bigl(\theta_i(q_i)\bigr)\) \\
MDH & \(\eta_{i,a}\) & \(\left|a_{i-1}\right|\) & \(\operatorname{sgn}_{+}(a_{i-1})\mathbf e_x\) & \(\mathbf R_{i-1}\) \\
MDH & \(\eta_{i,d}\) & \(\left|d_i(q_i)\right|\) & \(\operatorname{sgn}_{+}\!\left(d_i(q_i)\right)\mathbf e_z\) & \(\mathbf R_{i-1}\mathbf R_x(\alpha_{i-1})\) \\
\bottomrule
\end{tabularx}
\end{table}
The initialization \(\mathbf R_0\) ensures that every rotation prefix includes the fixed orientation of \(\mathcal F_0\) relative to the base frame. When \(\mathbf p_0\neq\mathbf 0_3\), the fixed base translation is included in \(\mathcal I\) through the configuration-independent item \(\eta_{\mathrm b}=(\|\mathbf p_0\|_2,\mathbf p_0/\|\mathbf p_0\|_2,\mathbf I_3)\).
With this definition, the translational forward kinematics take the unified form
\begin{equation}
\mathbf p_n(\mathbf q)= \hspace{-0.2cm}\sum_{k\in\mathcal I} \hspace{-0.1cm}\ell_k(\mathbf q)\mathbf u_k(\mathbf q) = \hspace{-0.2cm}\sum_{k\in\mathcal I} \hspace{-0.1cm} \ell_k(\mathbf q)\mathbf A_k(\mathbf q)\mathbf s_k(\mathbf q).
\label{eq:weighted_directional_sum}
\end{equation}
All coefficients $\ell_k(\mathbf q)$ are nonnegative, while the sign of each original DH translational coefficient is carried by $\mathbf s_k(\mathbf q)$ and consequently by $\mathbf u_k(\mathbf q)$. An item with $\ell_k(\mathbf q)=0$ contributes neither to the translational sum nor to the corresponding selector probability. The normalization factor and selector probabilities are introduced together with the circuit architectures. 
The terminal orientation is represented at the ideal expectation-value level through two propagated frame axes,
\begin{equation}
\mathbf x_n(\mathbf q)
=
\mathbf R_n(\mathbf q)\mathbf e_x,
\qquad
\mathbf z_n(\mathbf q)
=
\mathbf R_n(\mathbf q)\mathbf e_z.
\label{eq:terminal_frame_axes}
\end{equation}
The remaining axis follows from the right-handed frame constraint as $\mathbf y_n(\mathbf q) = \mathbf z_n(\mathbf q)\times\mathbf x_n(\mathbf q)$.
Under finite-shot execution, the corresponding estimated axes need not be exactly normalized or mutually orthogonal; their reconstruction and any subsequent projection onto \(SO(3)\) are specified in the experimental methodology.

\subsection{Single-Qubit States and Spatial Rotations}
\label{subsec:qubit_spatial_directions}

Let $\mathcal H_Q\simeq\mathbb C^2$ denote the Hilbert space of a single qubit and let $\mathcal D(\mathcal H_Q) = \left\{ \rho\in\mathbb C^{2\times 2} \mid \rho\succeq 0,\; \operatorname{Tr}(\rho)=1 \right\} $ denote the corresponding set of density operators. A pure state is represented by a rank-one density operator $\rho_\psi=|\psi\rangle\langle\psi|$, whereas a general $\rho\in\mathcal D(\mathcal H_Q)$ may be mixed \cite{Nielsen2010QuantumComputation}.
The state of a single qubit admits a complete real-valued representation through its Bloch vector. Let
\[
\sigma_x=
\begin{bmatrix}
0 & 1\\
1 & 0
\end{bmatrix},
\qquad
\sigma_y=
\begin{bmatrix}
0 & -\mathrm{i}\\
\mathrm{i} & 0
\end{bmatrix},
\qquad
\sigma_z=
\begin{bmatrix}
1 & 0\\
0 & -1
\end{bmatrix}
\]
be the Pauli operators, and let $\boldsymbol{\sigma} = (\sigma_x,\sigma_y,\sigma_z)$. Together with $I_2$, the Pauli operators form a basis for the Hermitian operators on $\mathcal H_Q$. Consequently, every single-qubit density operator can be written uniquely as
\begin{equation}
\rho
=
\frac{1}{2}
\left(
I_2+\mathbf r(\rho)\cdot\boldsymbol{\sigma}
\right),
\qquad
\mathbf r(\rho)
=
\begin{bmatrix}
\operatorname{Tr}(\rho\sigma_x)\\
\operatorname{Tr}(\rho\sigma_y)\\
\operatorname{Tr}(\rho\sigma_z)
\end{bmatrix}.
\label{eq:bloch_representation}
\end{equation}
The real vector $\mathbf r(\rho)\in\mathbb R^3$ is the Bloch vector of $\rho$, and its entries are the ideal expectation values of the three Pauli observables $r_\nu(\rho)=\langle\sigma_\nu\rangle_\rho = \operatorname{Tr}(\rho\sigma_\nu)$ for $\nu\in\{x,y,z\}$.
Physical single-qubit states satisfy $\|\mathbf r(\rho)\|_2\leq 1$, so the corresponding set $\mathbb B^3  = \left\{ \mathbf r\in\mathbb R^3 \mid \|\mathbf r\|_2\leq 1 \right\}$ is the Bloch ball. Pure states satisfy $\|\mathbf r(\rho)\|_2=1$ and lie on its boundary, the Bloch sphere $\mathbb S^2$, whereas mixed states lie in its interior. Thus, pure single-qubit states are in one-to-one correspondence with unit vectors in $\mathbb R^3$, up to the physically irrelevant global phase of their state-vector representatives. This three-dimensional representation applies specifically to individual qubits; a general multi-qubit state is not characterized by a single Bloch vector.
For $\nu\in\{x,y,z\}$ and $\phi\in\mathbb R$, define the single-qubit rotation
\[
U_\nu(\phi)
=
\exp\!\left(
-\mathrm{i}\frac{\phi}{2}\sigma_\nu
\right)
\in SU(2).
\]
Its adjoint action on a density operator induces the corresponding right-handed rotation of the Bloch vector
\begin{equation}
\mathbf r
\left(
U_\nu(\phi)\rho U_\nu^\dagger(\phi)
\right)
=
\mathbf R_\nu(\phi)\mathbf r(\rho),
\label{eq:bloch_induced_rotation}
\end{equation}
where $\mathbf R_\nu(\phi)\in SO(3)$ is the classical rotation matrix about the $\nu$-axis.
More generally, consider the ordered spatial rotation and a corresponding unitary representative
\[
\mathbf A
=
\mathbf R_{\nu_1}(\phi_1)
\cdots
\mathbf R_{\nu_m}(\phi_m)
\in SO(3)
\]
\[
U_{\mathbf A}
=
U_{\nu_1}(\phi_1)
\cdots
U_{\nu_m}(\phi_m)
\in SU(2).
\]
Repeated application of \eqref{eq:bloch_induced_rotation} gives
\begin{equation}
\mathbf r
\left(
U_{\mathbf A}\rho U_{\mathbf A}^{\dagger}
\right)
=
\mathbf A\,\mathbf r(\rho).
\label{eq:general_bloch_rotation}
\end{equation}
The rightmost transformation acts first in both ordered products. This ordering must be preserved because three-dimensional rotations generally do not commute.
The map from $SU(2)$ to $SO(3)$ is two-to-one. In particular, $U_{\mathbf A}$ and $-U_{\mathbf A}$ represent the same spatial rotation because $(-U_{\mathbf A})\rho(-U_{\mathbf A})^\dagger = U_{\mathbf A}\rho U_{\mathbf A}^\dagger$. 
The induced transformation of the Bloch vector is therefore independent of the choice between the two unitary representatives of the same element of $SO(3)$.

\subsection{Reduced States and Convex Aggregation of Qubit Observables}
\label{subsec:reduced_states}

Consider a composite quantum system with Hilbert space $\mathcal H_S\otimes\mathcal H_E\otimes\mathcal H_Q$, where \(S\) is a finite-dimensional label register, \(E\) collects any additional ancillary, item, or workspace subsystems, and \(Q\) is a single-qubit readout subsystem with $\mathcal H_Q\simeq\mathbb C^2$. Unless otherwise stated, this tensor-product order is adopted throughout the article.\\
Let $\rho_{SEQ}\in\mathcal D(\mathcal H_S\otimes\mathcal H_E\otimes\mathcal H_Q)$ be the joint state. If measurements are performed only on \(Q\), all locally accessible statistics are determined by its reduced density operator
\begin{equation}
\rho_Q
=
\operatorname{Tr}_{SE}\!\left(\rho_{SEQ}\right),
\label{eq:reduced_readout_state}
\end{equation}
where \(\operatorname{Tr}_{SE}\) denotes the partial trace over the unobserved subsystems \(S\) and \(E\). For any single-qubit observable \(O_Q=O_Q^\dagger\),
\begin{equation}
\operatorname{Tr}(\rho_Q O_Q)
=
\operatorname{Tr}
\left[
\rho_{SEQ}
\left(
I_S\otimes I_E\otimes O_Q
\right)
\right].
\label{eq:local_observable_reduced_state}
\end{equation}
Thus, \(\rho_Q\) contains all measurement statistics locally accessible from the readout qubit \cite{Nielsen2010QuantumComputation}.
The following abstract construction describes the reduced-state mechanism used later for weighted readout. Let \(\mathcal L\) be a finite set of branch labels and let \(K=|\mathcal L|\). A label register of \(s=\lceil\log_2 K\rceil\) qubits has Hilbert space $\mathcal H_S\simeq(\mathbb C^2)^{\otimes s}$ and provides at least \(K\) mutually orthogonal computational-basis states. Each active branch \(k\in\mathcal L\) is assigned a distinct basis state \(|k\rangle_S\), referred to as a label state. Computational-basis states not assigned to active branches, when present, are assumed to have zero amplitude. 
Let $\lambda_k\geq 0$ for every \(k\in\mathcal L\), with $\sum_{k\in\mathcal L}\lambda_k=1$. The label register may then be prepared in the normalized superposition $|\lambda\rangle_S = \sum_{k\in\mathcal L} \sqrt{\lambda_k}\,|k\rangle_S$, so that measurement of \(S\) in the computational basis returns label \(k\) with probability \(\lambda_k\).
For each branch \(k\), let \(|\psi_k\rangle_Q\) be the normalized state associated with the readout subsystem and let \(|\zeta_k\rangle_E\) be a normalized state collecting the branch-dependent contents of all other subsystems. After a generic label-controlled transformation, the composite system may have the form
\begin{equation}
|\Psi\rangle_{SEQ}
=
\sum_{k\in\mathcal L}
\sqrt{\lambda_k}\,
|k\rangle_S
|\zeta_k\rangle_E
|\psi_k\rangle_Q.
\label{eq:label_conditioned_joint_state}
\end{equation}
Thus, each selector branch correlates the label \(|k\rangle_S\) with a corresponding readout state \(|\psi_k\rangle_Q\) and, when required, with a branch-dependent state \(|\zeta_k\rangle_E\) of the remaining registers. The states \(|\zeta_k\rangle_E\) and \(|\psi_k\rangle_Q\) need not be mutually orthogonal. The state in \eqref{eq:label_conditioned_joint_state} is normalized because the label states are orthonormal, the branch states are normalized, and the coefficients \(\lambda_k\) sum to one. Depending on the branch states, the global state may be entangled across \(S\), \(E\), and \(Q\).
If only the readout subsystem is retained, tracing out \(S\) and \(E\) gives
\begin{equation}
\rho_Q
=
\operatorname{Tr}_{SE}
\left(
|\Psi\rangle\langle\Psi|
\right)
=
\sum_{k\in\mathcal L}
\lambda_k\rho_k,
\quad
\rho_k
=
|\psi_k\rangle\langle\psi_k|.
\label{eq:reduced_state_convex_mixture}
\end{equation}
The same reduced-state and observable-linearity relations hold when the branch states are mixed, with $\rho_k\in\mathcal D(\mathcal H_Q)$ replacing $|\psi_k\rangle\langle\psi_k|$. The pure-state construction is used here because it directly describes the ideal selector-conditioned circuits developed later.
The cross terms between distinct branches vanish because $\langle j|k\rangle_S=\delta_{jk}$. Consequently, the reduced state of \(Q\) is a convex mixture of the branch states, even when the global state in \eqref{eq:label_conditioned_joint_state} is coherent and entangled.
Expectation values are linear in the density operator. Therefore, for any single-qubit observable \(O_Q\),
\begin{equation}
\langle O_Q\rangle_{\rho_Q}
=
\sum_{k\in\mathcal L}
\lambda_k
\langle O_Q\rangle_{\rho_k}.
\label{eq:observable_convex_linearity}
\end{equation}
Applying \eqref{eq:observable_convex_linearity} to the three Pauli operators yields
\begin{equation}
\mathbf r(\rho_Q)
=
\sum_{k\in\mathcal L}
\lambda_k\mathbf r(\rho_k).
\label{eq:reduced_state_bloch_linearity}
\end{equation}
Hence, the Bloch vector of the readout qubit is the convex combination of the Bloch vectors associated with the branch states.
Even when all \(\rho_k\) are pure, the reduced state \(\rho_Q\) is generally mixed and satisfies $\left\| \mathbf r(\rho_Q) \right\|_2 \leq \sum_{k\in\mathcal L} \lambda_k \left\| \mathbf r(\rho_k) \right\|_2 \leq 1$.
The relation in \eqref{eq:reduced_state_bloch_linearity} describes a convex combination of density operators and their observable expectation values, rather than a coherent linear combination of the state vectors \(|\psi_k\rangle_Q\) within the readout subsystem. This distinction is fundamental when weighted spatial quantities are represented through the marginal state of a single qubit.

\subsection{Pauli Readout and Finite-Shot Estimation}
\label{subsec:pauli_readout}

For a single-qubit state \(\rho\), the Cartesian components of its Bloch vector are the ideal expectation values \(r_\nu(\rho)=\langle\sigma_\nu\rangle_\rho\). These quantities are not obtained deterministically from a single circuit execution. Instead, each expectation value is estimated from repeated preparations and projective measurements of the same quantum state.
A projective measurement of \(\sigma_\nu\) produces an outcome \(m_\nu\in\{-1,+1\}\) with probabilities $\Pr(m_\nu=\pm1) = \frac{1\pm r_\nu(\rho)}{2}$.
On gate-based quantum processors, measurement is conventionally performed in the computational basis. A \(\sigma_z\) measurement therefore requires no basis transformation, whereas \(\sigma_x\) and \(\sigma_y\) measurements are implemented by applying \(H\) and the sequence \(S^\dagger\) followed by \(H\), respectively, before computational-basis measurement.
Let \(N_\nu\in\mathbb N\) denote the number of shots allocated to the measurement setting associated with \(\sigma_\nu\), and let \(m_\nu^{(j)}\) be the outcome obtained in the \(j\)-th shot. The empirical estimator of the corresponding Bloch-vector component is
\begin{equation}
\widehat r_\nu^{(N_\nu)}
=
\frac{1}{N_\nu}
\sum_{j=1}^{N_\nu}
m_\nu^{(j)}
=
\frac{N_{\nu,+}-N_{\nu,-}}{N_\nu},
\label{eq:pauli_expectation_estimator}
\end{equation}
where \(N_{\nu,+}\) and \(N_{\nu,-}\) are the numbers of \(+1\) and \(-1\) outcomes, respectively. Throughout this article, a hat denotes a quantity estimated from a finite number of measurement shots, whereas the corresponding unhatted quantity denotes its ideal expectation-value counterpart.
Assuming independent and identically distributed shots, the estimator in \eqref{eq:pauli_expectation_estimator} is unbiased and satisfies $\mathbb E \left[ \widehat r_\nu^{(N_\nu)} \right] = r_\nu(\rho)$. Since \(m_\nu^2=1\), its variance is
\begin{equation}
\operatorname{Var}
\left[
\widehat r_\nu^{(N_\nu)}
\right]
=
\frac{1-r_\nu^2(\rho)}{N_\nu}
\leq
\frac{1}{N_\nu}.
\label{eq:pauli_shot_variance}
\end{equation}
The standard deviation therefore decreases as \(N_\nu^{-1/2}\), independently of the circuit used to prepare the state.
The three Pauli operators do not commute and cannot be jointly measured through a single projective measurement on one copy of a single-qubit state. A complete Bloch-vector estimate obtained from one readout qubit $\widehat{\mathbf r} = \left[ \widehat r_x^{(N_x)} \ \widehat r_y^{(N_y)} \ \widehat r_z^{(N_z)} \right]$ therefore requires three measurement settings,
with a total of \(N_x+N_y+N_z\) state preparations. Under an equal allocation of \(N_{\mathrm{shots}}\) shots per setting, the standard single-qubit protocol requires \(3N_{\mathrm{shots}}\) circuit executions. Alternatively, equivalent copies of the readout state may be prepared on distinct qubits and measured in different bases within the same circuit execution, exchanging additional quantum width and state-preparation operations for fewer measurement settings.
Because its components are estimated independently, the finite-shot vector \(\widehat{\mathbf r}\) is not guaranteed to satisfy \(\|\widehat{\mathbf r}\|_2\leq 1\), even when the ideal state is physical and pure. Similarly, independently estimated frame directions need not be exactly normalized or mutually orthogonal. Any normalization, physical-state projection, or geometric post-processing applied to these estimators must therefore be stated explicitly in the experimental methodology.

\section{Quantum Circuit Architectures for Robot Kinematics encoding}
\label{sec:main_contributions}

This section develops quantum circuit architectures for representing the forward kinematics of serial open-chain manipulators described through standard or modified Denavit--Hartenberg parameters. Starting from the directional decomposition established in Section~\ref{subsec:robot_model}, the proposed construction maps the DH-induced propagation of frame directions to parameterized single-qubit operations and realizes the resulting position and orientation quantities through alternative quantum readout strategies. We first define the common circuit-level target, then introduce the DH-derived directional-state preparation and terminal-frame orientation readout. We subsequently develop the compact IBSS and SCP position architectures and the NISQ-oriented IWR, tree-factorized, and parallel Cartesian variants.

\subsection{Problem Formulation and Architecture Overview}
\label{subsec:problem_formulation}

Consider the forward-kinematics map $\mathbf T_n(\mathbf q)$ defined in \eqref{eq:forward_kinematic_transform}, with end-effector position $\mathbf p_n(\mathbf q)$ and orientation $\mathbf R_n(\mathbf q)$ expressed in the base frame. The terminal-frame directions associated with $\mathbf R_n(\mathbf q)$ are defined in \eqref{eq:terminal_frame_axes}. The circuit-level objective is to associate each admissible configuration $\mathbf q\in\mathcal Q$ with parameterized quantum operations whose measurable outputs reproduce these kinematic quantities while preserving the ordered dependence on the underlying DH transformations.
The compact pose interface comprises a position-readout qubit $Q_p$ and two orientation-readout qubits $Q_x$ and $Q_z$, with $\mathcal H_{Q_p}\simeq\mathcal H_{Q_x}\simeq\mathcal H_{Q_z}\simeq\mathbb C^2$. Let $\mathcal C$ denote the complete circuit register, let $\rho_{\mathcal C}^{(0)}$ be its initial state, and let
\[
\rho_{\mathcal C}(\mathbf q)
=
U_{\mathcal E}(\mathbf q)
\rho_{\mathcal C}^{(0)}
U_{\mathcal E}^{\dagger}(\mathbf q)
\]
be the ideal circuit output. Whenever a designated readout qubit $Q_\lambda$, with $\lambda\in\{p,x,z\}$, is present, its reduced state is $\rho_\lambda(\mathbf q) = \operatorname{Tr}_{\mathcal C\setminus Q_\lambda} \left[ \rho_{\mathcal C}(\mathbf q) \right]$.
At the ideal expectation-value level, the compact pose interface is required to satisfy
\begin{equation}
\begin{aligned}
\mathbf r\!\left(\rho_p(\mathbf q)\right)
&=
\frac{\mathbf p_n(\mathbf q)}{W(\mathbf q)},
\\
\mathbf r\!\left(\rho_x(\mathbf q)\right)
=
\mathbf x_n(\mathbf q) &,\quad 
\mathbf r\!\left(\rho_z(\mathbf q)\right)
=
\mathbf z_n(\mathbf q),
\end{aligned}
\label{eq:compact_pose_target}
\end{equation}
where $W(\mathbf q)>0$ is the classical metric scale induced by the active translational contributions and is defined together with the position-encoding construction below. The normalized position target satisfies $\left\| \frac{\mathbf p_n(\mathbf q)}{W(\mathbf q)} \right\|_2 \leq 1$ and can therefore be represented by a valid single-qubit Bloch vector. The metric position is reconstructed as
$\mathbf p_n(\mathbf q) = W(\mathbf q)\, \mathbf r\!\left(\rho_p(\mathbf q)\right)$.
The two orientation readouts determine the complete terminal frame according to the relations established in \eqref{eq:terminal_frame_axes}.

The IBSS and SCP architectures developed below realize the compact position condition in \eqref{eq:compact_pose_target} through different selector-controlled aggregation mechanisms. They are complemented by three NISQ-oriented alternatives. IWR removes compact quantum aggregation and reconstructs the same position from separately measured directional contributions; tree-factorized variants reorganize selector preparation and controlled operations; and parallel Cartesian readout exchanges additional qubits and replicated state preparation for fewer measurement settings. These architectures expose different trade-offs among circuit width, controlled-operation depth, routing overhead, measurement settings, repeated circuit executions, and classical post-processing.

\subsection{DH-Derived Directional Qubit Preparation}
\label{subsec:directional_quantum_encoding}

Consider a translational item $\eta_k(\mathbf q)=(\ell_k(\mathbf q),\mathbf s_k(\mathbf q),\mathbf A_k(\mathbf q))$, with $k\in\mathcal I$, as defined in Section~\ref{subsec:robot_model}. Its Cartesian contribution is $\ell_k(\mathbf q)\mathbf u_k(\mathbf q)$, where $\ell_k(\mathbf q)>0$ is the metric magnitude and $\mathbf u_k(\mathbf q)=\mathbf A_k(\mathbf q)\mathbf s_k(\mathbf q)\in\mathbb S^2$ is the propagated signed direction. The purpose of the circuit construction is to prepare a single-qubit state whose Bloch vector reproduces $\mathbf u_k(\mathbf q)$. The positive coefficient $\ell_k(\mathbf q)$ remains classical and is subsequently used to determine the relative weight of the corresponding contribution in the position-aggregation architectures. The construction defines a directional-state preparation independently of the register on which it is instantiated.
Let $\mathcal H_Q\simeq\mathbb C^2$ denote the Hilbert space of the qubit on which a directional state is prepared, and let $\rho^{(0)}=|0\rangle\langle0|$ be the reference state, with $\mathbf r(\rho^{(0)})=\mathbf e_z$. For each active item, the preparation consists of a seed unitary $P_k(\mathbf q)\in SU(2)$ followed by a propagation unitary $U_k(\mathbf q)\in SU(2)$. Their composition defines
\begin{equation}
V_k(\mathbf q)=U_k(\mathbf q)P_k(\mathbf q),
\qquad
V_k:\mathcal Q\rightarrow SU(2),
\label{eq:item_preparation_unitary}
\end{equation}
and the corresponding directional state is
\begin{equation}
\rho_k(\mathbf q)
=
V_k(\mathbf q)\rho^{(0)}V_k^\dagger(\mathbf q).
\label{eq:propagated_item_state}
\end{equation}
The seed unitary is chosen such that $\mathbf r\!\left(P_k(\mathbf q)\rho^{(0)}P_k^\dagger(\mathbf q)\right)=\mathbf s_k(\mathbf q)$. For the DH-derived translational items in Table~\ref{tab:dh_candidate_items}, $\mathbf s_k(\mathbf q)\in\{\pm\mathbf e_x,\pm\mathbf e_z\}$ because the sign of the original DH coefficient has already been absorbed into the local translation axis. A convenient set of canonical seed preparations is $P_{+z}=I_2$, $P_{-z}=U_y(\pi)$, $P_{+x}=U_y(\pi/2)$, and $P_{-x}=U_y(-\pi/2)$. Accordingly, $P_k(\mathbf q)$ is selected from these four operations according to $\mathbf s_k(\mathbf q)$. For items whose underlying DH coefficient has a configuration-dependent sign, such as a prismatic displacement that may cross zero within its admissible range, the corresponding seed preparation changes consistently with that sign. Configurations for which the coefficient is zero generate no active item by definition of $\mathcal I$.
For the fixed item $k$, write the corresponding rotation prefix as a product of elementary Cartesian rotations,
\[
\mathbf A_k(\mathbf q)
=
\mathbf R_{\nu_1}\!\left(\phi_1(\mathbf q)\right)
\cdots
\mathbf R_{\nu_m}\!\left(\phi_m(\mathbf q)\right),
\]
where $m\in\mathbb N_0$, $\nu_j\in\{x,y,z\}$, and $\phi_j:\mathcal Q\to\mathbb R$, for $j=1,\ldots,m$. The dependence of $m$, $\nu_j$, and $\phi_j$ on the selected item $k$ is left implicit to avoid overloading the notation. When $m=0$, the product is the identity.
The propagation unitary is obtained by replacing each spatial rotation with its corresponding single-qubit $SU(2)$ representative while preserving the same ordered product,
\[
U_k(\mathbf q)
=
U_{\nu_1}\!\left(\phi_1(\mathbf q)\right)
\cdots
U_{\nu_m}\!\left(\phi_m(\mathbf q)\right),
\]
with $U_k(\mathbf q)=I_2$ when $m=0$. In both products, the rightmost transformation acts first, so the generally noncommutative ordering of the classical DH rotation prefix is preserved exactly.
The rotation axes and angles are inherited directly from the fixed base orientation and from the SDH or MDH rotation sequence associated with item $k$. Fixed geometric parameters generate configuration-independent rotations, whereas revolute coordinates enter through the same functions $\theta_i(q_i)$ appearing in the classical kinematic model. Prismatic coordinates do not modify the rotation prefix but can affect both the positive metric magnitude $\ell_k(\mathbf q)$ and the signed seed direction $\mathbf s_k(\mathbf q)$.
Using the seed-preparation condition and the $SU(2)$-to-$SO(3)$ correspondence established in \eqref{eq:general_bloch_rotation}, the resulting state satisfies
\begin{equation}
\begin{aligned}
\mathbf r\!\left(\rho_k(\mathbf q)\right)
&=
\mathbf r\!\left(
U_k(\mathbf q)P_k(\mathbf q)
\rho^{(0)}
P_k^\dagger(\mathbf q)U_k^\dagger(\mathbf q)
\right)\\
&=
\mathbf A_k(\mathbf q)
\mathbf r\!\left(
P_k(\mathbf q)\rho^{(0)}P_k^\dagger(\mathbf q)
\right)\\
&=
\mathbf A_k(\mathbf q)\mathbf s_k(\mathbf q)
=
\mathbf u_k(\mathbf q).
\end{aligned}
\label{eq:exact_directional_encoding}
\end{equation}
Hence, at the ideal expectation-value level, the three Pauli observables of the prepared qubit reproduce the Cartesian components of the corresponding signed DH-propagated unit direction expressed in the base frame. Together with the positive classical coefficient $\ell_k(\mathbf q)$, this state represents the complete translational contribution $\ell_k(\mathbf q)\mathbf u_k(\mathbf q)$ appearing in \eqref{eq:weighted_directional_sum}.
The state in \eqref{eq:propagated_item_state} is independent of the choice between the two $SU(2)$ representatives of the same spatial rotation. It is also independent of the particular elementary-gate decomposition used to synthesize $U_k(\mathbf q)$, provided that the implemented unitary induces the required adjoint action $\mathbf A_k(\mathbf q)$ on the Bloch vector. When the fixed base-translation item defined in Section~\ref{subsec:robot_model} is present and its direction is not aligned with a canonical Cartesian axis, its seed can analogously be prepared by any single-qubit unitary whose Bloch action maps $\mathbf e_z$ to $\mathbf p_0/\|\mathbf p_0\|_2$.

\begin{figure}[!t]
\centering
\begin{minipage}[t]{0.45\columnwidth}
    \centering
    \includegraphics[width=\linewidth]{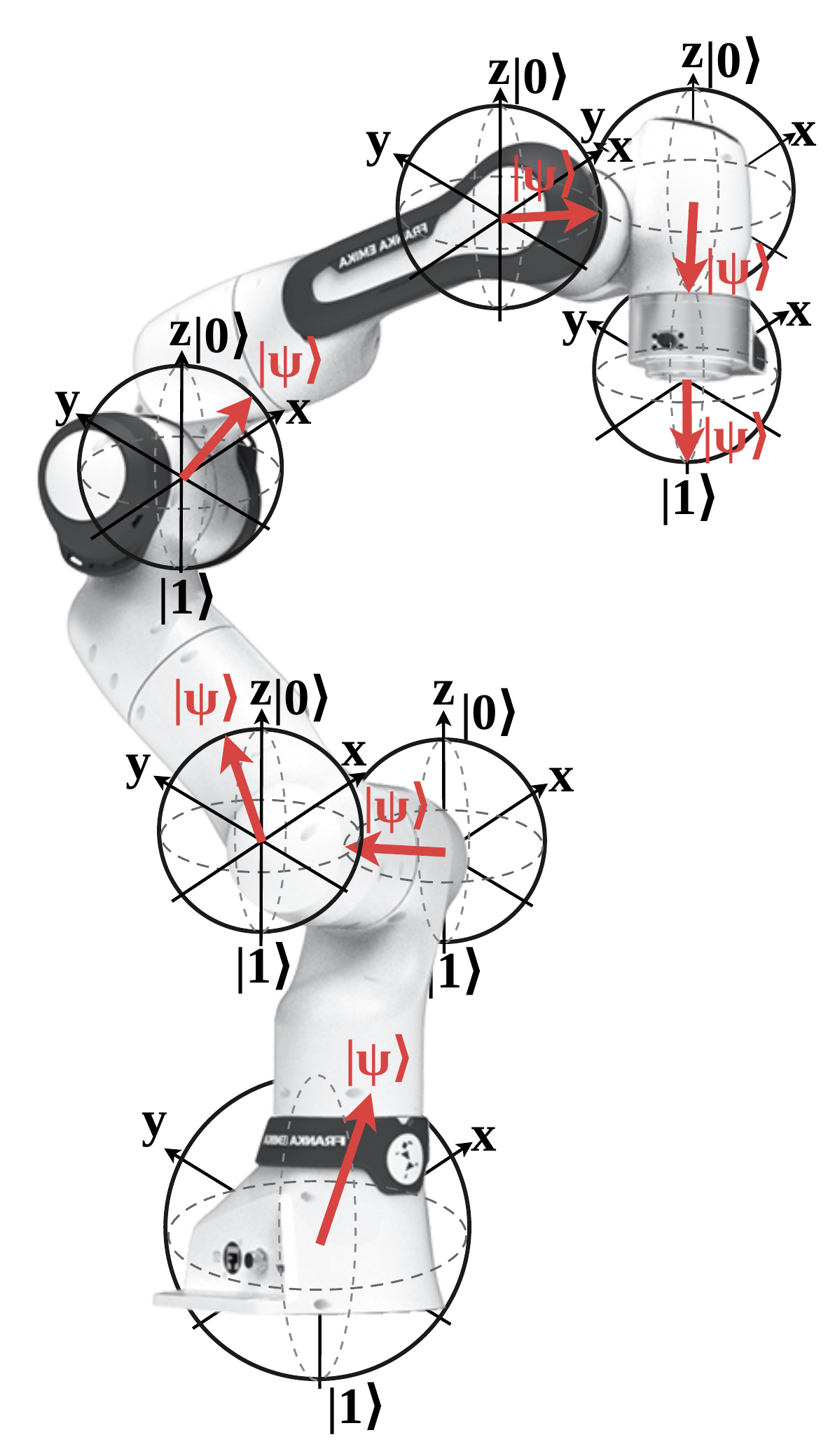}
    
    \vspace{1mm}
    \textbf{(a) Link contribution}
\end{minipage}
\hfill
\begin{minipage}[t]{0.54\columnwidth}
    \centering
    \includegraphics[width=\linewidth]{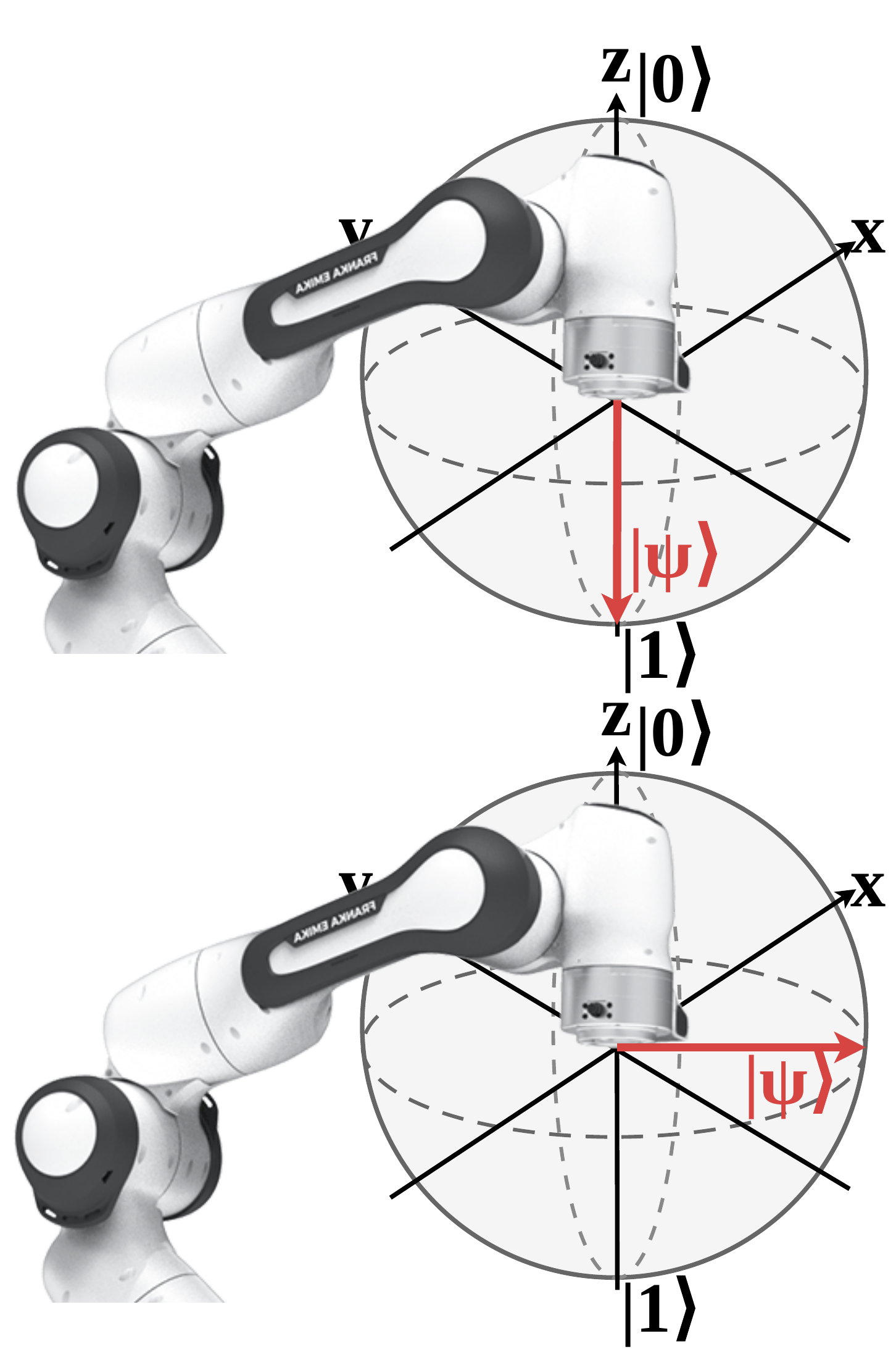}
    
    \vspace{1mm}
    \textbf{(b) Terminal-frame axes}
\end{minipage}
\caption{Examples of DH\-derived directional-state preparation for the FR3 manipulator. (a) Item-level encoding of a translational contribution, where the signed propagated direction is represented by the Bloch vector of a single-qubit state, while the corresponding positive metric magnitude remains classical. (b) Directional encoding of the terminal-frame axes used for orientation readout.}
\label{fig:fr3_directional_encoding}
\end{figure}

The same directional-state preparation applies directly to the terminal-frame orientation. As defined in \eqref{eq:terminal_frame_axes}, the terminal rotation is characterized by the propagated axes $\mathbf x_n(\mathbf q)=\mathbf R_n(\mathbf q)\mathbf e_x$ and $\mathbf z_n(\mathbf q)=\mathbf R_n(\mathbf q)\mathbf e_z$. Let $U_n(\mathbf q)$ denote the single-qubit unitary obtained from the complete ordered DH rotation sequence defining $\mathbf R_n(\mathbf q)$, according to the same $SO(3)$-to-$SU(2)$ construction used for $U_k(\mathbf q)$. Two orientation-readout qubits, $Q_x$ and $Q_z$, are initialized from $\rho^{(0)}$ using $P_x=U_y(\pi/2)$ and $P_z=I_2$, respectively, and propagated through the same $U_n(\mathbf q)$. Their states are therefore
\[
\rho_\xi(\mathbf q)
=
U_n(\mathbf q)P_\xi\rho^{(0)}P_\xi^\dagger U_n^\dagger(\mathbf q),
\qquad
\xi\in\{x,z\},
\]
and satisfy, at the ideal expectation-value level,
$\mathbf r\!\left(\rho_x(\mathbf q)\right)=\mathbf R_n(\mathbf q)\mathbf e_x=\mathbf x_n(\mathbf q)$ and
$\mathbf r\!\left(\rho_z(\mathbf q)\right)=\mathbf R_n(\mathbf q)\mathbf e_z=\mathbf z_n(\mathbf q)$.
Using the right-handed-frame reconstruction defined in \eqref{eq:terminal_frame_axes}, the complete terminal orientation is  recovered as
\[
\mathbf R_n(\mathbf q)
=
\begin{bmatrix}
\mathbf r\!\left(\rho_x(\mathbf q)\right) &
\mathbf r\!\left(\rho_z(\mathbf q)\right)\times\mathbf r\!\left(\rho_x(\mathbf q)\right) &
\mathbf r\!\left(\rho_z(\mathbf q)\right)
\end{bmatrix}.
\]
Thus, two instances of the directional-state preparation are sufficient to represent the complete end-effector orientation in the ideal model. This orientation branch is independent of the translational readout architecture and can therefore be combined unchanged with each of the position-encoding constructions introduced next.

\section{Item-Wise Readout}
\label{subsec:iwr}

The Item-Wise Readout (IWR) architecture represents the translational component of the kinematic model through a register of independently prepared directional qubits and reconstructs the end-effector position from their measured Pauli expectation values. Let $Q_{\mathcal I}=\{Q_k\}_{k\in\mathcal I}$ denote the item register, containing one qubit for each of the $K=|\mathcal I|$ translational items induced by the DH model. Each item state $\rho_k(\mathbf q)$ is prepared according to Section~\ref{subsec:directional_quantum_encoding} and, under the ideal construction, satisfies
\(
\mathbf r\!\left(\rho_k(\mathbf q)\right)=\mathbf u_k(\mathbf q),
\)
where $\mathbf u_k(\mathbf q)$ is the unit directional contribution associated with the corresponding metric coefficient $\ell_k(\mathbf q)$.
Since the directional states are pure in the ideal representation, each state may be written as $\rho_k(\mathbf q)=|\psi_k(\mathbf q)\rangle\langle\psi_k(\mathbf q)|$, with $|\psi_k(\mathbf q)\rangle=V_k(\mathbf q)|0\rangle$. The complete translational register is therefore prepared as the product state
\[
|\Psi_{\mathcal I}(\mathbf q)\rangle
=
\bigotimes_{k\in\mathcal I}
|\psi_k(\mathbf q)\rangle_{Q_k}.
\]
The coefficients $\ell_k(\mathbf q)$ remain classical quantities and preserve the metric information that is not contained in the normalized Bloch-vector encoding.
For each item qubit, the Cartesian components of $\mathbf u_k(\mathbf q)$ are obtained from the Pauli expectation values
\(
\mathbf r\!\left(\rho_k(\mathbf q)\right)
=
\begin{bmatrix}
\langle X_k\rangle &
\langle Y_k\rangle &
\langle Z_k\rangle
\end{bmatrix}^{\mathsf T}.
\)
The translational output is then reconstructed by classical weighted aggregation of the item-level readouts,
\begin{equation}
\mathbf p_n^{\mathrm{IWR}}(\mathbf q)
=
\sum_{k\in\mathcal I}
\ell_k(\mathbf q)\,
\mathbf r\!\left(\rho_k(\mathbf q)\right)
=
\mathbf p_n(\mathbf q).
\label{eq:iwr_position_reconstruction}
\end{equation}
The final equality follows directly from the exact directional encoding and the weighted directional decomposition in \eqref{eq:weighted_directional_sum}. Hence, the exact end-effector position is recovered at the ideal expectation-value level without requiring the translational sum itself to be represented by an additional quantum subsystem.
Using the normalization factor $W(\mathbf q)$ and the normalized weights $\lambda_k(\mathbf q)=\ell_k(\mathbf q)/W(\mathbf q)$ introduced above, the same reconstruction can equivalently be expressed as
\[
\frac{\mathbf p_n^{\mathrm{IWR}}(\mathbf q)}{W(\mathbf q)}
=
\sum_{k\in\mathcal I}
\lambda_k(\mathbf q)\,
\mathbf r\!\left(\rho_k(\mathbf q)\right)
=
\frac{\mathbf p_n(\mathbf q)}{W(\mathbf q)}.
\]
This form separates the normalized directional information extracted from the quantum register from the classical scale factor $W(\mathbf q)$ required to recover the position in physical units.
If $\ell_k(\mathbf q)=0$ for a particular configuration, the corresponding item contributes zero to \eqref{eq:iwr_position_reconstruction}. The register structure therefore remains fixed over the configuration space, while configuration-dependent metric coefficients determine the contribution of each encoded direction to the final Cartesian position.
The orientation component is obtained independently from the two orientation-readout qubits $Q_x$ and $Q_z$ introduced in Section~\ref{subsec:directional_quantum_encoding}. Their Bloch vectors encode two axes of the terminal frame and provide the information required to reconstruct the end-effector orientation according to the procedure defined above. The complete IWR pose representation therefore consists of the translational item register $Q_{\mathcal I}$ together with the orientation-readout qubits $Q_x$ and $Q_z$. Quantum processing prepares and measures the directional states, whereas classical post-processing combines the measured translational directions with the corresponding metric coefficients and reconstructs the terminal-frame orientation.
\begin{figure}[!t]
\centering
\includegraphics[width=\columnwidth]{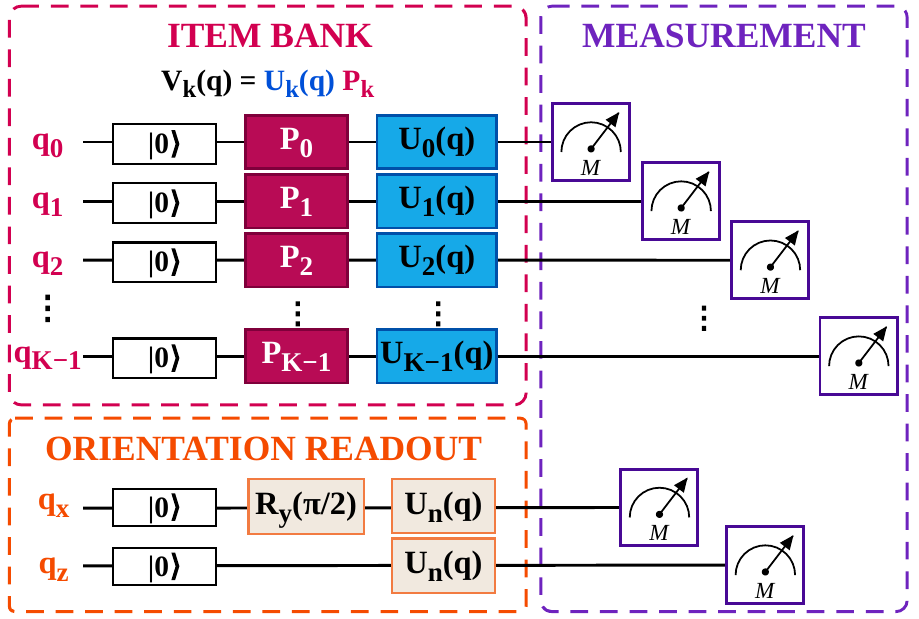}
\caption{Item-wise readout (IWR) architecture. A register of $K$ item qubits encodes the DH-derived translational directions, while $Q_x$ and $Q_z$ encode the terminal-frame orientation. The end-effector position is reconstructed classically from the measured item-level Bloch vectors and their corresponding metric coefficients.}
\label{fig:iwr_architecture}
\end{figure}
The IWR architecture consequently implements the structure-preserving quantum kinematic representation using independently prepared directional qubits, local Pauli measurements, and classical weighted reconstruction. This organization avoids inter-item quantum operations in the translational readout stage and exposes a direct correspondence between DH-derived kinematic contributions and measurable quantum states.

\bibliographystyle{IEEEtran}
\bibliography{bibliography}

\EOD

\end{document}